\documentclass[10pt,twocolumn,letterpaper]{article}
\usepackage{multirow}
\usepackage[justification=centering]{caption}
\usepackage{subfigure}
\usepackage{subcaption}
\usepackage{gensymb}
\usepackage[table]{xcolor}
\usepackage[accsupp]{axessibility}

\clearpage
\usepackage{cvpr}              % To produce the CAMERA-READY version
\definecolor{cvprblue}{rgb}{0.21,0.49,0.74}
\usepackage[pagebackref,breaklinks,colorlinks,citecolor=cvprblue]{hyperref}

\def\paperID{3} % *** Enter the Paper ID here
\def\confName{CVPR}
\def\confYear{2024}

\title{UrbanSARFloods: Sentinel-1 SLC-Based Benchmark Dataset for Urban and Open-Area Flood Mapping}

\author{Jie Zhao, Zhitong Xiong, Xiao Xiang Zhu\\
Department of Aerospace and Geodesy, Data Science in Earth Observation,\\
Technical University of Munich, Arcisstraße 21, 80333 Munich, Germany\\
{\tt\small \{jie.zhao, zhitong.xiong,xiaoxiang.zhu\}@tum.de}
}

\begin{document}
\maketitle
\begin{abstract}

Due to its cloud-penetrating capability and independence from solar illumination, satellite Synthetic Aperture Radar (SAR) is the preferred data source for large-scale flood mapping, providing global coverage and including various land cover classes. However, most studies on large-scale SAR-derived flood mapping using deep learning algorithms have primarily focused on flooded open areas, utilizing available open-access datasets (e.g., Sen1Floods11) and with limited attention to urban floods. To address this gap, we introduce \textbf{UrbanSARFloods}, a floodwater dataset featuring pre-processed Sentinel-1 intensity data and interferometric coherence imagery acquired before and during flood events. It contains 8,879 $512\times 512$ chips covering 807,500 $km^2$ across 20 land cover classes and 5 continents, spanning 18 flood events. We used UrbanSARFloods to benchmark existing state-of-the-art convolutional neural networks (CNNs) for segmenting open and urban flood areas. Our findings indicate that prevalent approaches, including the Weighted Cross-Entropy (WCE) loss and the application of transfer learning with pretrained models, fall short in overcoming the obstacles posed by imbalanced data and the constraints of a small training dataset. Urban flood detection remains challenging. Future research should explore strategies for addressing imbalanced data challenges and investigate transfer learning's potential for SAR-based large-scale flood mapping. Besides, expanding this dataset to include additional flood events holds promise for enhancing its utility and contributing to advancements in flood mapping techniques. The UrbanSARFloods dataset, including training, validation data, and raw data, can be found at ~\url{https://github.com/jie666-6/UrbanSARFloods}.

\end{abstract}    
\section{Introduction}
\label{sec:intro}

%Flood mapping especially urban flood mapping is demanded

\begin{figure*}[h]
    \centering
    \includegraphics[width=0.86\linewidth]{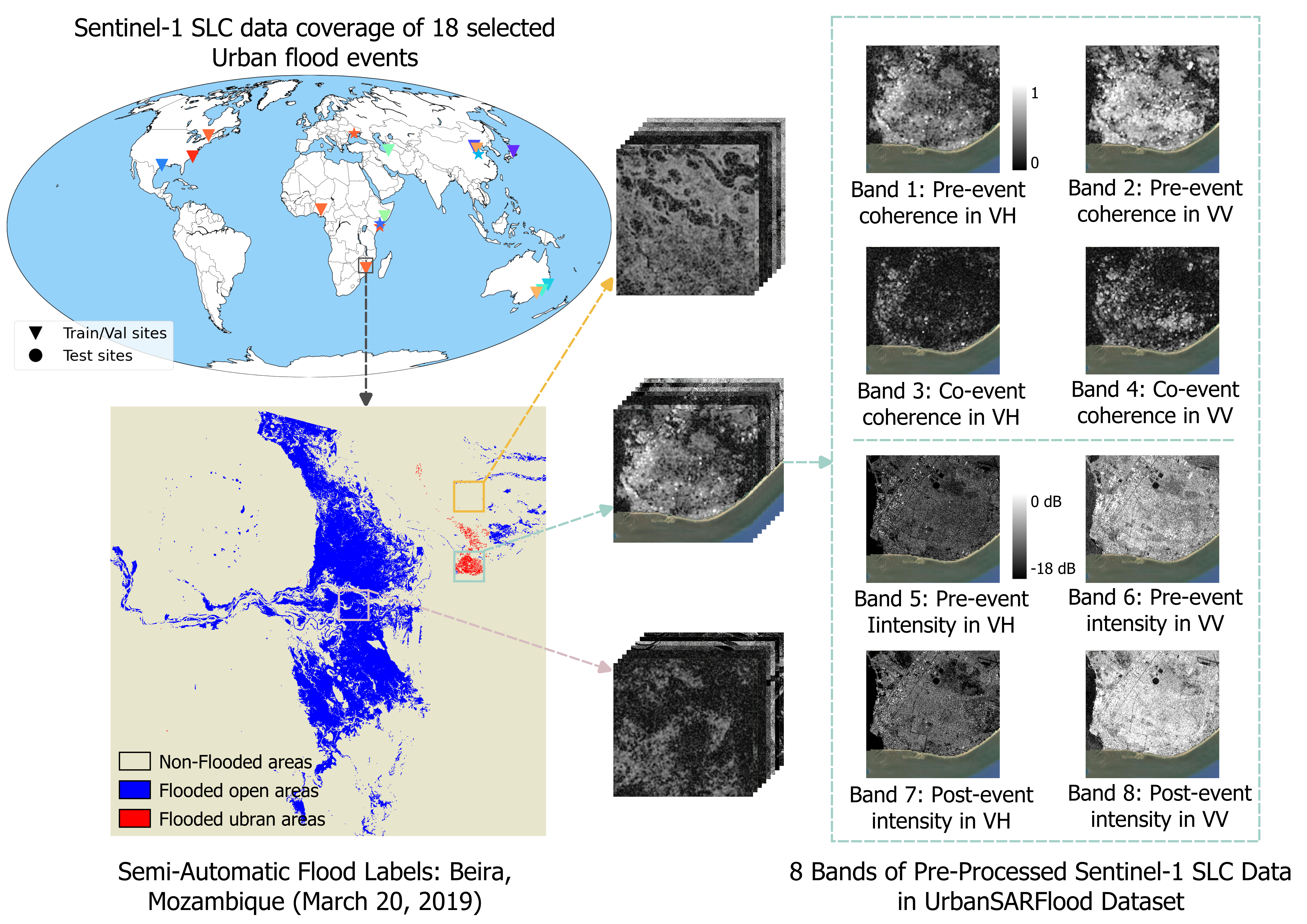}
    \caption{Overview of the UrbanSARFloods dataset.}
    \label{fig:Dataset}
\end{figure*}

As one of the most devastating natural disasters worldwide, floods have impacted billions of people~\citep{rentschler2022flood}. Additionally, it has been reported that the frequency and severity of flooding have increased due to intensified heavy precipitation patterns~\citep{Caretta2022water}. Thus, there is a growing demand for flood mapping, monitoring, and forecasting on a global scale, not only for high-profile catastrophic flooding events but also for those unreported floods. Within this context, Earth Observation (EO) data, particularly satellite EO data, plays a critical role in understanding and quantifying the extent and depths of flooding events on a large scale, thus presenting a significant opportunity for global flood mapping and monitoring. Currently, a variety of satellite optical data sources such as MODIS, VIIRS, Landsat constellations, Sentinel-2, and PlanetScope, as well as satellite synthetic aperture radar (SAR) data including Sentinel-1, COSMO-SkyMed, Radarsat Constellation, TerraSAR-X, and GaoFen-3, are widely employed for flood mapping. However, satellite optical data can be easily contaminated by clouds although they offer longer temporal coverage and are easier to interpret. Therefore, satellite SAR data is preferred due to its ability to be acquired regardless of weather conditions (i.e., clouds) and solar illumination. Furthermore, it is worth noting that the biggest difference between large-scale flood mapping and region-scale flood mapping is that the methods should identify floodwater in various land cover classes, including bare soils, sparsely vegetated areas, urban areas, agricultural fields, etc. However, to the best of our knowledge, most large-scale SAR-derived flood mapping studies mainly focus on flooded open areas, i.e., bare soils and sparsely vegetated areas~\citep[e.g.,][]{jiang2021rapid,zhao2021large,yang2021high,bauer2022satellite,li2023robust}, while large-scale SAR-derived flood mapping considering both open areas and urban areas has not been well investigated yet. To clarify, in this study, the term "\textit{large-scale urban flood mapping}" is used to denote flood mapping that covers both open areas and urban areas.

Recently, a growing number of studies~\cite[e.g.][]{chini2019sentinel,iervolino2015flooding,baghermanesh2022urban,gokon2023detecting} have utilized SAR data, showing its efficacy in identifying floodwater in urban areas. It has been found that both SAR intensity and Interferometric SAR (InSAR) coherence are critical in urban flood mapping~\cite{li2019urban,zhao2022urban}. Specifically, in SAR data, the totally submerged areas (such as bare soils and buildings) typically exhibit lower backscatter than the background because the SAR signal has been reflected away from the sensor due to the specular reflection in the calm open water surface. Partially submerged buildings, on the other hand, often show relatively higher backscatter due to the double-bounce effects between the water surface and the buildings' facades. Additionally, the coherence of flooded buildings experiences a sharp drop with the appearance of floodwater, serving as a useful indicator in flood detection, particularly when changes in the intensity of flooded buildings are too small to be detected. Furthermore, it is essential to consider different polarizations simultaneously, as the backscatter is influenced by both the orientation of the buildings and the line-of-sight (LOS) of the sensor~\cite{pelich2022mapping, zhao2022urban}. However, since most existing methods have not been tested on large datasets in both temporal and spatial dimensions, their generalizability and robustness in large-scale urban flood mapping remain unknown.

Although deep learning (DL) techniques have a transformative impact on the remote sensing field, their integration into SAR-based large-scale urban flood mapping applications remains limited. Only a handful of studies~\cite{zhao2022urban,yang2023promoting,li2019urban} have ventured into this domain so far. Comparing their generalizability and robustness is challenging due to variations in data, flood events, and data preprocessing methodologies employed across studies. Meanwhile, we have observed that a growing number of DL-based large-scale flood mapping studies~\cite[e.g.][]{mateo2021towards,konapala2021exploring,katiyar2021near,li2023assessment,yadav2024unsupervised,garg2023cross} were carried out thanks to the publication of the georeferenced Sen1Floods11 dataset~\cite{bonafilia2020sen1floods11}. This phenomenon of public datasets driving advances across various research fields has also been demonstrated in different domains. For instance, in the field of image classification and object detection in computer vision, ImageNet has provided researchers worldwide with access to vast amounts of data, while its test dataset has served to document the enhancements in computer vision capabilities~\citep{bonafilia2020sen1floods11}. Therefore, it is believed that one of the bottlenecks hindering the advancement of DL-based methodologies in the field of SAR-based large-scale urban flood mapping is the absence of a benchmark dataset.

% bottleneck of AI in urban flood mapping is benchmark dataset
To bridge this gap and engage more AI researchers in advancing large-scale urban flood mapping with SAR data, we developed a georeferenced flood dataset named \textbf{UrbanSARFloods}. The dataset comprises flooded urban and open areas, with SAR intensity and InSAR coherence data obtained pre- and post-event in VV and VH polarizations using Sentinel-1 SLC data(\cref{fig:Dataset}). It offers global coverage, consisting of 8,879 chips of 512$\times$512 20m pixels across 18 flood events, spanning 807,500 $km^2$. All 8,879 chips underwent semi-automatic labeling via conventional remote sensing methods. Moreover, three subsets from selected events were manually annotated using high-resolution optical data (i.e., 3m PlanetScope and 10cm UAV optical RGB orthophoto). Semi-automatic and manual labels are utilized during testing. Further details are provided in Section~\ref{sec:dataset}. Leveraging the UrbanSARFloods dataset, we evaluate flood detection performance of various semantic segmentation models, including both pre-trained and scratch-trained ones. Our findings highlight the persistent challenge in urban flood detection, primarily due to data imbalance and limited dataset availability.

\section{Related Work}
\label{sec:SOTA}

\begin{table*}[]
\caption{Summary of current publicly available flood-related datasets.}
\label{tab:flood_dataset}
\fontsize{8}{10}\selectfont

\begin{tabular}{c|c|c|c|c}
\hline
Dataset                           & Platform                    & Modality                                     & Resolution {[}m{]}           & Flood Labels                        \\ \hline
MM-Flood~\cite{bprf-jf62-22}                         & Spaceborne                  & SAR intensity,   DEM, hydrography maps       &  10                            & Flooded open   areas                \\ \hline
Sen1Floods11~\cite{bonafilia2020sen1floods11}                      & Spaceborne                  & SAR   intensity, Multispectral               & 10                           & Flooded open areas                  \\ \hline
Sen12-Flood~\cite{w6xz-s898-20}                       & Spaceborne                  & SAR   intensity, Multispectral               & 10                           & Flooded open areas                  \\ \hline
% ETCI-2021~\cite{nasa_etci2021}                         & Spaceborne                  & SAR   intensity                              & 20                           & Flooded open areas                  \\ \hline
\multirow{3}{*}{RAPID-NRT~\cite{yang2021high}}        & \multirow{3}{*}{Spaceborne} & SAR intensity, topography,                   & \multirow{3}{*}{10}          & \multirow{3}{*}{Flooded open areas} \\
                                  &                             & water occurrence, land cover classification, &                              &                                     \\
                                  &                             & river width, hydrography, and water type     &                              &                                     \\ \hline
% S1GFloods~\cite{saleh2023dam}                         & Spaceborne                  & SAR   intensity                              & -                            & Flooded open areas                  \\ \hline
\multirow{2}{*}{SpaceNet 6~\cite{shermeyer2020spacenet}}         & \multirow{2}{*}{Spaceborne} & \multirow{2}{*}{Optical data}                & \multirow{2}{*}{0.3   - 0.8} & Flooded building /                  \\
                                  &                             &                                              &                              & Flooded road                        \\ \hline
\multirow{2}{*}{EU Flood Dataset~\cite{barz2019enhancing}} & Airborne /                  & \multirow{2}{*}{Multispectral   data}        & \multirow{2}{*}{-}           & \multirow{2}{*}{Flooding}           \\
                                  & Ground collected data       &                                              &                              &                                     \\ \hline
Hurricane Harvey Floods~\cite{github_repo}           & Spaceborne                  & SAR intensity, InSAR coherence, Optical data & 1,   2, 10                   & Flood  Damage building              \\ \hline
Flood Area Segmentation~\cite{Flood_Area_Segmentation_dataset}           & Airborne                    & Optical   data                               & -                            & Flood                               \\ \hline
Roadway flooding image~\cite{sazara_dataset}    & Ground   collected data     & Optical   data                               & -                            & Flood                               \\ \hline
ML4Floods~\cite{julia_global_2023}                         & Spaceborne                  & Optical   data                               & 10 - 30                            & Flooded open areas                  \\ \hline
\multirow{2}{*}{FloodNet~\cite{9460988}}         & \multirow{2}{*}{Airborne}   & \multirow{2}{*}{Optical   data}              & \multirow{2}{*}{-}           & Flooded building /                  \\
                                  &                             &                                              &                              & Flooded road                        \\ \hline
California flood dataset~\cite{8798991}          & Spaceborne                  & SAR   intensity, Multispectral               & -                            & Flooded open areas                  \\ \hline
\end{tabular}
\end{table*}

\subsection{Flood-related benchmark datasets}

According to several remote sensing data platforms, such as EarthNet (\url{https://earthnets.github.io/})~\cite{xiong2022earthnets}, the IEEE platform (\url{https://eod-grss-ieee.com/dataset-search}), and SpaceML from Frontier Development Lab (\url{https://spaceml.org/repo}), there are several flood-related datasets available. As shown in~\cref{tab:flood_dataset}, it is evident that most flood-related datasets currently focus on flooded open areas. However, only three datasets (SpaceNet, Hurricane Harvey Floods, FloodNet) provide annotations related to flood-affected buildings. However, the Hurricane Harvey Floods dataset is designed for flood-damaged buildings instead of flooded buildings, while the SpaceNet and FloodNet datasets provide annotations for two flood classes (i.e., flooded buildings and flooded roads) using high-resolution optical data. Besides, another recent S1GFloods dataset, which reportedly includes wetlands, riverine areas, mountainous regions, urban and rural areas, and vegetation~\cite{saleh2023dam}, annotated flooded areas using Sentinel-1 intensity data only. Therefore, only severely flooded buildings may be included, while most of the flooded built-up areas cannot be effectively included, as it has been established that InSAR coherence is important complementary information with SAR intensity when it comes to detecting flooded buildings using SAR data.

Indeed, there is currently no suitable SAR dataset specifically designed for large-scale urban flood mapping. While it is possible to gather flood labels from various studies, as exemplified by~\cite{brill2021extrapolating} who shared their manually annotated flood labels as a supplementary file, inconsistencies in annotation granularity stemming from variations in spatial resolution and labeling styles may compromise the reliability of conclusions drawn from such data. Additionally, it has been observed that certain non-flooded areas, such as tarmac and shrubland, exhibit characteristics similar to flooded areas, potentially leading to overestimations in large-scale flood mapping efforts~\cite{zhao2021deriving}. In comparison to existing datasets, UrbanSARFloods offers a distinctive advantage by encompassing two flooded classes (i.e., flooded open areas and flooded urban areas) characterized by SAR intensity and InSAR coherence data. This unique feature renders UrbanSARFloods particularly suitable for large-scale urban flood mapping applications.

\subsection{Semantic Segmentation in flood mapping application}

Currently, numerous semantic segmentation methods have recently been applied to flood mapping applications using SAR data. For instance, \cite{nemni2020fully} evaluated the performance of UNet~\cite{ronneberger2015u}, XNet~\cite{bullock2019xnet} and UNet with ResNet as the backbone on the UNOSAT Flood Dataset (\url{https://unosat-rm.cern.ch/FloodAI/apps/MMR/}), stating that the performance of the models did not significantly differ. However, the UNet with ResNet as the backbone was reported as the most favorable due to its greater flexibility in the choice of precision/recall tradeoff. Subsequently, \cite{katiyar2021near} indicated that UNet outperformed SegNet~\cite{badrinarayanan2017segnet} using the Sen1Floods11 dataset in flood mapping, attributed to its skip connection architecture. Furthermore, \cite{lv2022high} proposed a modified DeepLabV3+ model employing MobileNetv2 as the backbone for detecting flooded open areas using a C-band commercial SAR satellite data Hisea-1, which is surpassed SegNet, UNet, and DeepLabv3+~\cite{chen2018encoder} in both accuracy and inference time~\cite{lv2022high}. 

Recently, \cite{jamali2024residual} introduced the Residual Wave Vision U-Net (WVResU-Net), trained and tested on Sentinel-1 data, integrating advanced Vision Multi-Layer Perceptrons (MLPs) and ResU-Net, exhibiting significant superiority over several well-known CNN and ViT DL models. Beside the supervising learning, an unsupervised CNN model has been introduced in SAR-based flood mapping~\cite{jiang2021rapid}, where SAR images are pre-segmented using the graph-based Felzenszwalb and Huttenlocher (Felz) segmentation algorithm, followed by the CNN being employed to generate the final flood map. In recent research, \cite{yadav2024unsupervised} introduced a lightweight unsupervised flood mapping DL model, named Contrastive ConvLSTM Variational AutoEncoder (CLVAE), which employs fully self-supervised training with simplified contrastive learning techniques. However, it should be noted that all the above-mentioned studies were developed for flooded open areas instead of flooded urban areas using SAR data. To the best of our knowledge, only a handful of studies \cite{zhao2022urban, li2019urban, yang2023promoting} have been conducted specifically focusing on SAR-based urban flood mapping using DL techniques, where many popular semantic segmentation models such as UNet++ and Deeplabv3+ have not been investigated.

\section{Datasets}
\label{sec:dataset}

Our dataset contains 18 Sentinel-1-covered urban flood events, in which the changes caused by floodwater in open areas and urban areas could be measured. The geographic distribution of selected flood events is shown in~\cref{fig:Dataset}, while more detailed information is listed in~\cref{tab:summary_event}. The imagery in our dataset has 8 bands, including 2 bands (VV and VH) for Sentinel-1 intensity acquired pre-event, 2 bands (VV and VH) for Sentinel-1 intensity acquired post-event, 2 bands (VV and VH) for Sentinel-1 coherence acquired pre-event, and 2 bands (VV and VH) for Sentinel-1 intensity acquired co-event.

\begin{table*}[]
\centering
\caption{Detailed description of global flood events in UrbanSARFloods dataset. Events labeled with * signify the inclusion of optical data.}
\label{tab:summary_event}
\fontsize{8}{10}\selectfont
\begin{tabular}{c|c|c|c|c|c|ccc}
\hline
\multirow{2}{*}{Continent} &   \multirow{2}{*}{Location} &  \multirow{2}{*}{Event Date} &  \multirow{2}{*}{Image Size} &  \multirow{2}{*}{Absolute Orbit} &  \multirow{2}{*}{Path} &  \multicolumn{3}{c}{Number of tiles} \\ \cline{7-9} 
 &   &   &   &   &   &  \multicolumn{1}{c}{NF} &  \multicolumn{1}{c}{FO} &  FU \\ \hline
\multirow{3}{*}{North America} &  Houston, US &  30 August 2017 &  14918 × 12981 &  7169 &  143 &  \multicolumn{1}{c|}{274} &  \multicolumn{1}{c|}{323} &  129 \\  
                    \cline{2-9}  &  Lumberton, US &  11 Oct 2016 &  9931 × 6465 &  13449 &  77 &  \multicolumn{1}{c|}{147} &  \multicolumn{1}{c|}{201} &  2 \\ 
                    \cline{2-9}  &  Sainte-Marthe-sur-le-Lac, Canada &  02 May 2019 &  21638 × 11184 &  27055 &  33 &  \multicolumn{1}{c|}{279} &  \multicolumn{1}{c|}{431} &  2 \\ \hline
\multirow{5}{*}{Africa} &  Beledweyne, Somalia &  08 May 2018 &  14293 × 11656 &  21807 &  35 &  \multicolumn{1}{c|}{495} &  \multicolumn{1}{c|}{73} &  4 \\ 
            \cline{2-9}  &  Beira, Mozambique &  20 March 2019 &  14904 × 12608 &  15432 &  6 &  \multicolumn{1}{c|}{133} &  \multicolumn{1}{c|}{89} &  14 \\ 
            \cline{2-9}  &  Beledweyne, Somalia &  14 Nov 2023 &  15457 × 12634 &  51207 &  35 &  \multicolumn{1}{c|}{455} &  \multicolumn{1}{c|}{56} &  20 \\ 
            \cline{2-9}  &  Jubba, Somalia* &  01 Dec 2023 &  \begin{tabular}[c]{@{}c@{}}15548 × 13078   \\ 15454 × 12710\end{tabular} &  \begin{tabular}[c]{@{}c@{}}50580   \\ 51455\end{tabular} &  \begin{tabular}[c]{@{}c@{}}108\\ 108\end{tabular} &  \multicolumn{1}{c|}{\begin{tabular}[c]{@{}c@{}}400\\ 460\end{tabular}} &  \multicolumn{1}{c|}{\begin{tabular}[c]{@{}c@{}}59\\ 65\end{tabular}} &  \begin{tabular}[c]{@{}c@{}}13\\ 10\end{tabular} \\ 
            \cline{2-9}  &  Lokoja, Niger &  13 Oct 2022 &  15500 × 12587 &  44902 &  30 &  \multicolumn{1}{c|}{427} &  \multicolumn{1}{c|}{107} &  1 \\ \hline
\multirow{5}{*}{Asia} &  Iwaki/Koriyama, Japan &  12 Oct 2019 &  11751 ×   10096 &  18447 &  46 &  \multicolumn{1}{c|}{353} &  \multicolumn{1}{c|}{158} &  19 \\ \cline{2-9} 
                     &  Weihui, China* &  27 July 2021 &  18927 × 12245 &  38962 &  40 &  \multicolumn{1}{c|}{293} &  \multicolumn{1}{c|}{221} &  135 \\ \cline{2-9} 
                     &  Aqqala, Iran &   29 March 2021 &  19549 ×   12580 &  26554 &  57 &  \multicolumn{1}{c|}{333} &  \multicolumn{1}{c|}{351} &  25 \\ \cline{2-9} 
                    &  Zhuozhou, China &  05 August 2023 &  19906 ×   12207 &  49739 &  142 &  \multicolumn{1}{c|}{332} &  \multicolumn{1}{c|}{204} &  137 \\ \cline{2-9} 
                    &  Langfang, China &  05 August 2023 &  19458 ×   12220 &  49739 &  142 &  \multicolumn{1}{c|}{279} &  \multicolumn{1}{c|}{269} &  117 \\ \cline{2-9} \hline
                   %  &  Sunamganj, Myanmar &  23 June 2022 &   &  43797 &  150 &  \multicolumn{1}{c|}{} &  \multicolumn{1}{c|}{} &   \\ \hline
\multirow{3}{*}{Oceania} &  Coraki, Australia &  2 March 2022 &  18160 × 13358 &  42146 &  74 &  \multicolumn{1}{c|}{105} &  \multicolumn{1}{c|}{54} &  11 \\ \cline{2-9} 
                        &  Sydney, Australia &  24 March 2021 &  19495 ×   13582 &  37144 &  147 &  \multicolumn{1}{c|}{468} &  \multicolumn{1}{c|}{95} &  6 \\ \cline{2-9} 
                        &  Port Macquarie, Australia & 19 March 2021 &  18774 × 13460 &  37071 &  74 &  \multicolumn{1}{c|}{124} &  \multicolumn{1}{c|}{89} &  26 \\ \hline
Europe &  NovaKakhovka, Ukraine* &  09 June 2023 &  22596 × 12226 &  48911 &  14 &  \multicolumn{1}{c|}{103} &  \multicolumn{1}{c|}{612} &  37 \\ \hline
\end{tabular}
\end{table*}

\subsection{Preprocessing}

The Sentinel-1 Level-1 Interferometric Wide Swath SLC data, downloaded from \href{https://asf.alaska.edu/datasets/daac/sentinel-1/}{Alaska Satellite Facility (ASF)}, were used to extract interferograms and then calibrated and transformed into intensity (in dB). The multilooking (4 looks in the range and 1 looks in the azimuth) was carried out in order to get the interferogram with the square pixels. A Goldstein filter with a size of 9 $\times$ 9 pixels was applied to the interferogram to reduce noise in the phase, and then the interferometric coherence was estimated by a moving window of 9 $\times$ 9 pixels. All the intensity and coherence data were geocoded to World Geodetic System (WGS) 1984 longitude and latitude with 20 m spatial resolution. 

\subsection{Label generation}

\subsubsection{Semi-automatic Labeling}
\label{sec:annotation}
In this study, generating annotations covering the entire Sentinel-1 image frame can only be achieved through conventional remote sensing approaches. This is because no high-resolution optical images acquired on the same acquisition dates as Sentinel-1 data are capable of covering the entire image frame ($\sim$43,000 $km^2$). Three-step annotation is needed: 1) the open flooded areas are extracted by applying a hierarchical Split-based change detection approach (i.e., HSBA) \citep{chini2017hierarchical} to SAR intensity imagery; 2) the urban floods are extracted using a threshold (fixed at 0.3 based on trial-and-error) on difference interferometric coherence image (i.e., pre-event coherence - co-event coherence). The built-up areas and non-built-up areas are distinguished using World Settlement Footprint 2019 (WSF2019) \citep{marconcini2021understanding}; 3) isolated objects with a small number of pixels are eliminated, whose threshold for removal was determined on a case-by-case basis by remote sensing analysts. It should be noted that for both flood classes, once a pixel has been annotated as flooded in at least one polarization, it is annotated as flooded.

In the annotation data, flooded open areas are designated with a pixel value of 1, flood urban pixels are assigned a value of 2, and non-flooded regions are represented by a pixel value of 0.

\subsubsection{Hand Labeling}
For the areas where high-resolution optical data is available, trained remote sensing analysts annotated open floods and urban floods using centimeter-level UAV-based RGB imagery. All annotations were carried out using QGIS software. Due to the spatial resolution difference between RGB imagery and Sentinel-1 data, analysts need to remove some small flooded areas that cannot be recognized in Sentinel-1 data due to its coarse resolution.

\subsection{Training, Validation and Testing Data}
We aim to achieve diversity in addressing real-world flood issues, thereby advancing the development of effective flood detection methodologies. Hence, we consider not only the quantity of images but also their representativeness across various flood scenarios during data division. Specifically, we aim for training and validation datasets that encompass floods occurring in various land cover classes and different environmental conditions, while the testing datasets should feature floods from diverse locations, allowing us to evaluate the robustness and transferability of different flood detection methods. Therefore, three flood events located in Africa, Asia, and Europe were selected as testing cases due to the availability of high-resolution optical data, while the remaining 15 flood events were used for model training and validation.

For the 15 flood events utilized for training and validation, all imagery was partitioned into 512$\times$512 pixel non-overlapping chips. In our dataset, we retained all chips where no flood exists, as it is believed that the features of non-flood pixels also contribute to the improved detection of flooded pixels. More specifically, concerning large-scale flood mapping, many non-flooded areas exhibit flood-lookalike characteristics, such as the sparse shrubland near Beledeweyne, Somalia, which may lead to confusion for flood detection models and result in over-detection~\citep{zhao2021deriving} if no reliable annotation data from those specific areas is involved. Moreover, flood pixels typically constitute only a small fraction of the entire scene, even during catastrophic flood events, especially when dealing with 20m spatial resolution SAR data covering an area of 43,000 $km^2$. Consequently, flood datasets often exhibit an extreme data imbalance ratio ($\rho > 1000$), especially in urban flood cases. Therefore, the training and validation data division was carried out using stratified sampling strategies based on flood event cases.

Firstly, all tiles were classified into Non-flooded ($NF$) tiles, Flooded open areas ($FO$) tiles, and Flooded urban ($FU$) tiles by examining the maximal value in the annotation data generated in~\cref{sec:annotation}: if the maximal value in the annotation data of a tile is 2, then the tile is classified as an $FU$ tile; if the maximal value is 1, then the tile is classified as an $FO$ tile; otherwise, the tile is classified as an $NF$ tile. Then, both the $FO$ and $FU$ tiles were further divided into two subclasses separately: tiles with $\rho > 1000$ were assigned as subclass 1, while tiles with $\rho < 1000$ were assigned as subclass 2. In other words, all tiles were classified into 5 categories: $NF$, $FO_1$, $FO_2$, $FU_1$, and $FU_2$. For each category, 70\% of the tiles were allocated to the training dataset, while the remaining tiles were utilized for the validation dataset. 

Following this scheme, our dataset comprises 8,879 non-overlapping chips covering 807,500 $km^2$: 2,408 from three selected study sites for testing only; 4,501 for training, and 1,970 for validation across the remaining 15 events.

\subsection{Statistics of the UrbanSARFloods Dataset}

\begin{figure}[h]
    \centering
    \includegraphics[width=0.86\linewidth]{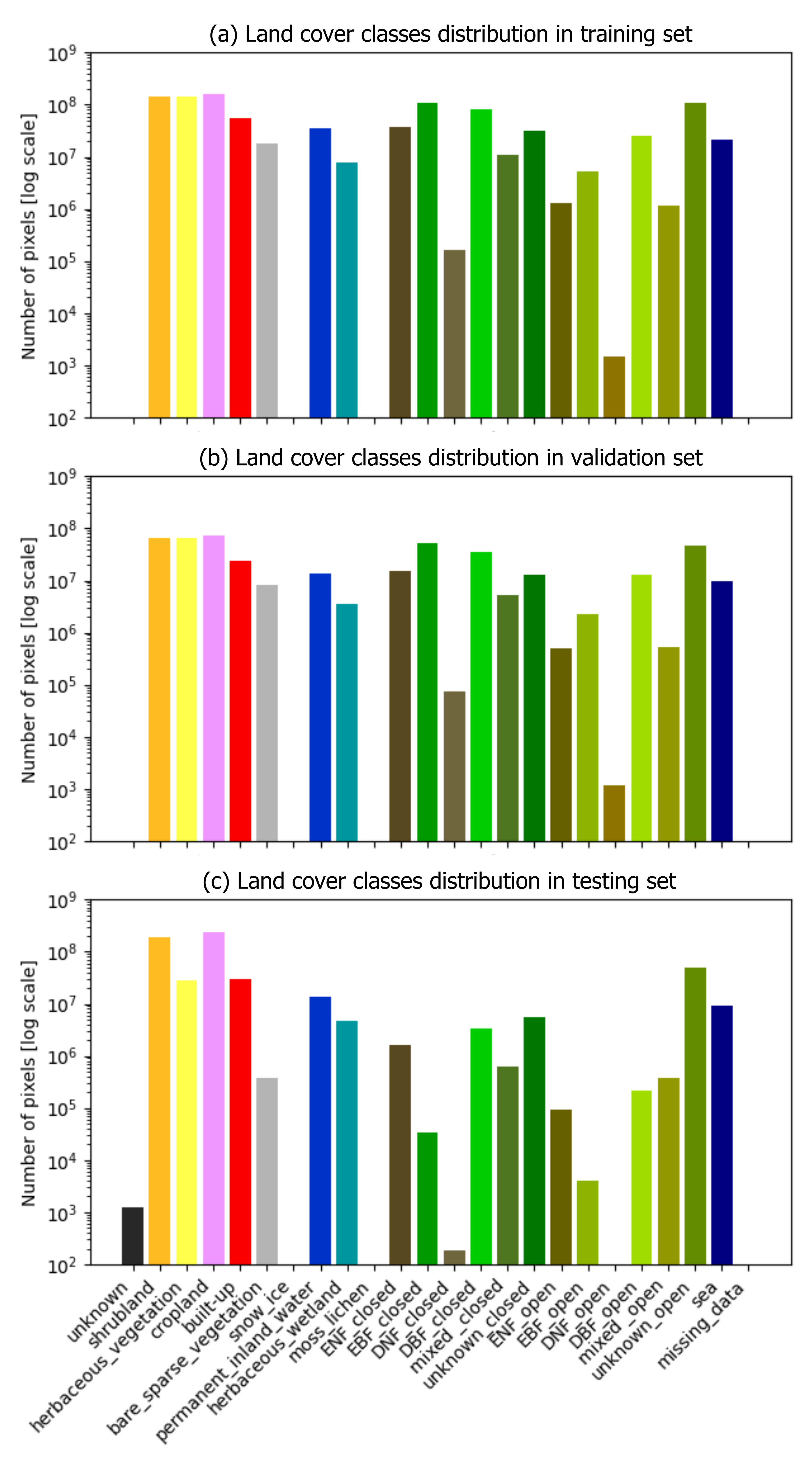}
    \caption{Statistics of label distribution of UrbanSARFloods.}
    \label{fig:analysis_LCM}
\end{figure}

\begin{figure}[]
 \centering
 \begin{subfigure}
     \centering
     \includegraphics[width=0.13\textwidth]{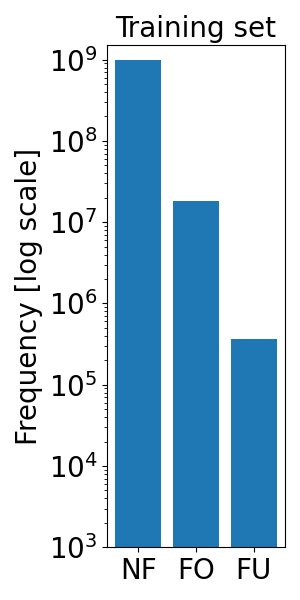}
     % \label{fig:train}
 \end{subfigure}
 \hspace{1em}
 \begin{subfigure}
     \centering
     \includegraphics[width=0.13\textwidth]{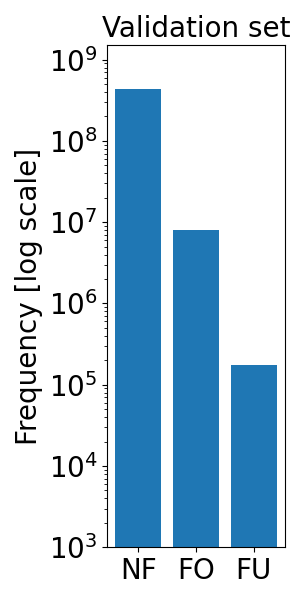}
     % \label{fig:valid}
 \end{subfigure}
 \hspace{1em}
 \begin{subfigure}
     \centering
     \includegraphics[width=0.13\textwidth]{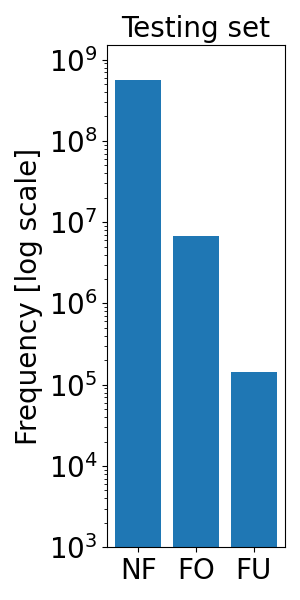}
     % \label{fig:valid}
 \end{subfigure}
 
 \hfill
    \caption{Statistics of semi-automatic label distribution of UrbanSARFloods.}
    \label{fig:analysis_GT}
\end{figure}

To have a better understanding of the dataset, we also show the statistics of the land cover classes distribution in training, validation, and testing dataset separately, where the Copernicus Global Land cover map 2019~\citep{buchhorn2020copernicus} was involved for such analysis. \cref{fig:analysis_LCM} demonstrates that both the training set, validation set, and testing set contain different land cover classes and the distribution of land cover classes is similar. Also, we display the label distributions of the training set, validation set, and testing set, where the annotation obeys clear long-tailed distributions and indicates the serious data imbalance issue (\cref{fig:analysis_GT}).

% \begin{figure}[!]
%  \centering
%  \begin{subfigure}
%      \centering
%      \includegraphics[width=0.5\textwidth]{figures/LCM2019_hist_Train_6.png}
%      % \label{fig:train}
%  \end{subfigure}
%  \hfill
%  \begin{subfigure}
%      \centering
%      \includegraphics[width=0.5\textwidth]{figures/LCM2019_hist_Valid_6.png}
%      % \label{fig:valid}
%  \end{subfigure}
%  \hfill
%     \caption{Statistics of land cover information of UrbanSARFloods.}
%     \label{fig:analysis_LCM}
% \end{figure}

\section{Experiment and results}

In this dataset, we have explicitly provided the official splits for the training, validation, and test subsets. Offering a well-defined dataset and official split is crucial for ensuring reproducibility. In addition, the different semantic segmentation models were trained on UrbanSARFloods dataset, including Unet~\citep{ronneberger2015u}, Unet++~\citep{zhou2018unet++}, MANet~\citep{fan2020ma}, Linknet~\citep{chaurasia2017linknet}, FPN~\citep{lin2017feature}, PSPNet~\citep{zhao2017pyramid}, PAN~\citep{li2018pyramid}, DeepLabV3~\citep{chen2017rethinking} and DeeplabV3+~\citep{chen2018encoder}. 

\subsection{Implementation details}

% All models are implemented using PyTorch and executed on an NVIDIA A40 GPU. The public codebase provided by Segmentation Models PyTorch~\cite{Iakubovskii:2019} was employed for this purpose. Given our aim to establish straightforward baselines for the datasets rather than optimize for the best possible models, we did not conduct an exhaustive hyperparameter search. Training for all models was conducted for 100 epochs. The input images, initially 512$\times$512 in size, were randomly cropped into 256$\times$256 dimensions with random horizontal and vertical flips and random rotation (i.e., 90$\degree$, 180$\degree$, 270$\degree$) applied for data augmentation. The batch sizes were set to 12, while we utilized the Adam optimizer with an initial learning rate of 1e-5 and a weight decay coefficient of 1e-4. The Weighted Cross-Entropy (WCE) loss was employed to address the class imbalance among the three classes. In order to evaluate the efficacy of transfer learning in flood mapping, each model underwent two training regimens: one in which the model's weights were initialized via Xavier initialization, and another in which the model was pre-trained using the ImageNet dataset. For all the experiments, the Precision, Recall, and F1 score of each class are used as the evaluation metrics. 

All models were implemented using PyTorch and executed on an NVIDIA A40 GPU. The public codebase provided by Segmentation Models PyTorch~\cite{Iakubovskii:2019} was employed for this purpose. Given our aim to establish straightforward baselines for the datasets rather than optimize for the best possible models, we did not conduct an exhaustive hyperparameter search. Training for all models was conducted for 100 epochs. The input images, initially sized 512$\times$512, were randomly cropped into 256$\times$256 dimensions with random horizontal and vertical flips, as well as random rotations (i.e., 90$\degree$, 180$\degree$, 270$\degree$), applied for data augmentation. The batch sizes were set to 12, while we utilized the Adam optimizer with an initial learning rate of 1e-5 and a weight decay coefficient of 1e-4. The Weighted Cross-Entropy (WCE) loss was employed to address the class imbalance among the three classes. To evaluate the efficacy of transfer learning in flood mapping, each model underwent two training regimens: one in which the model's weights were initialized via Xavier initialization, and another in which the model was pre-trained using the ImageNet dataset. For all experiments, Precision, Recall, and F1 score of each class were used as the evaluation metrics.

\subsection{Evaluation of State-of-the-art semantic segmentation models}

\begin{figure}[h]
    \centering
    \includegraphics[width=0.86\linewidth]{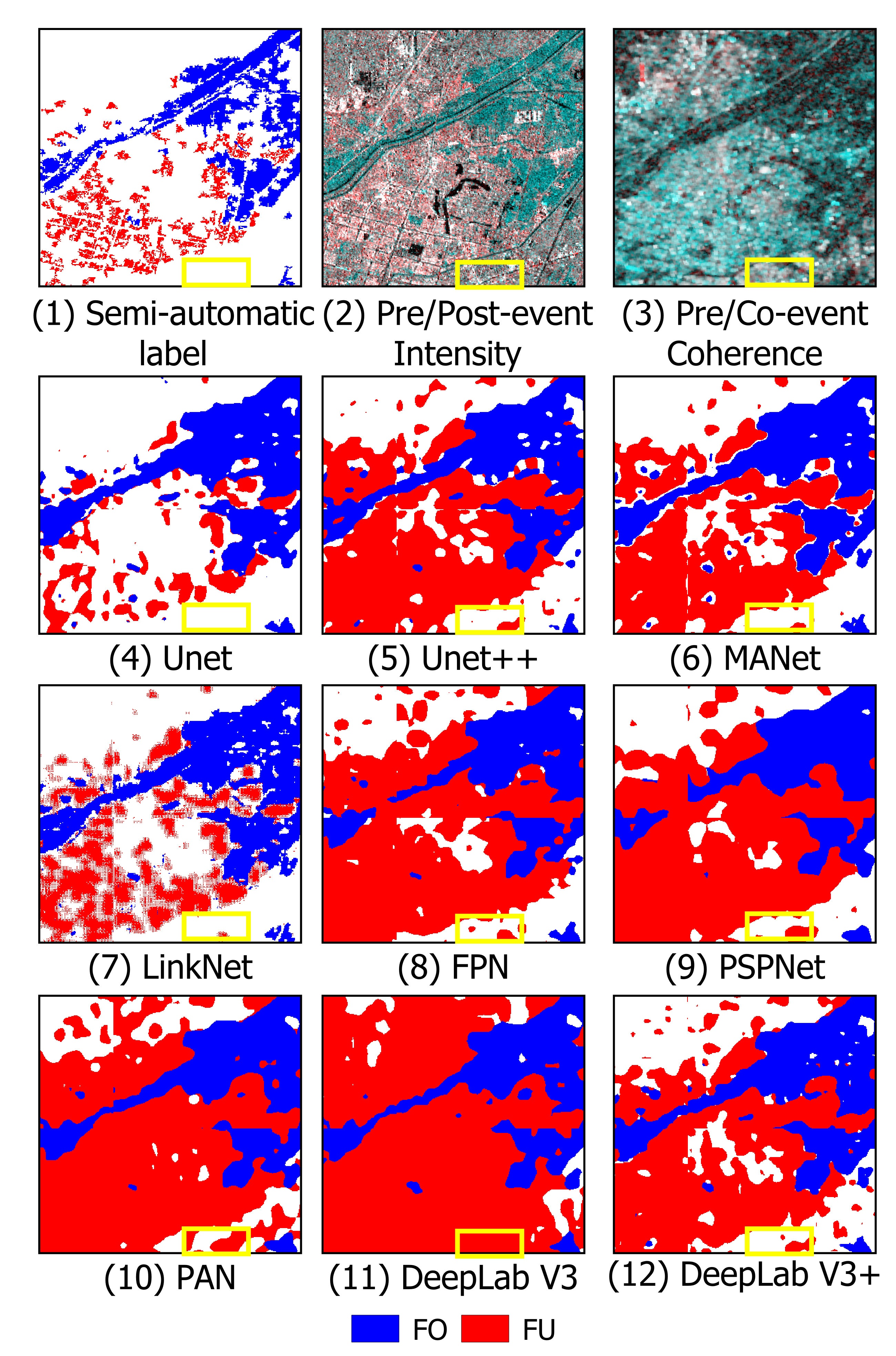}
    \caption{Example of one test site (Weihui): flood label data, SAR data, and generated flood maps using different models trained from scratch.}
    \label{fig:weihui}
\end{figure}

We evaluated nine existing semantic segmentation models available in Segmentation Models PyTorch~\cite{Iakubovskii:2019}. The results of two flood classes are presented in~\cref{tab:results_evaluation_large_FO} and~\cref{tab:results_evaluation_large_FU}, separately. As is shown in~\cref{tab:results_evaluation_large_FO}, the F1 score for FO in Weihui and Jubba ranges from 0.51 to 0.77 due to relatively low precision, indicating overestimation of FO. An example of Weihui is shown in~\cref{fig:weihui}, where FO is displayed in blue and FU is displayed in red. Within the RGB combination of pre-event/post-event intensity, FO is cyan and FU is red in~\cref{fig:weihui} (2). Similarly, within the RGB combination of pre-event and co-event coherence, FU is cyan in~\cref{fig:weihui} (3). Thus, combining the label data in~\cref{fig:weihui} (1) and SAR data in~\cref{fig:weihui} (2-3), it is clear that most overestimation of FO mainly exists in the boundary of FO. However, the F1 source of FO is much lower in the NovaKakhovka with the corresponding precision below 0.2. Compared with the 3m Planetscope data acquired on the same date, there are two sources of those FO false alarms: non-flooded agriculture fields and wind-affected permanent water surfaces having similar characterises of FO. 

When it comes to the quantitative results of FU, the F1 scores and their corresponding precision are lower than 0.1 (\cref{tab:results_evaluation_large_FU}), indicating too many false alarms of FU exist. Besides the difficulty in distinguishing flooded urban pixels from other FU-lookalike pixels, some pixels that definitely do not have FU-lookalike features have been wrongly classified as FU, as is shown in the yellow box in~\cref{fig:weihui}. In addition, there is a disparity in performance between FO and FU, which can be attributed to the imbalanced data issue, despite our efforts to address it using the Weighted Cross-Entropy (WCE) loss.

Furthermore, we conducted an evaluation of all flood maps using manually annotated data, and no significant differences were observed. Therefore, it is inferred that the relatively poor performance of all models is attributable to the challenges inherent in large-scale flood mapping applications, rather than being solely attributed to the semi-automatic label data itself.

\subsection{Evaluation of transfer learning}

Furthermore, all models pretrained on ImageNet were fine-tuned and tested on our dataset. Results are presented in Tables~\ref{tab:results_evaluation_large_FO_pretrain} and~\ref{tab:results_evaluation_large_FU_pretrain}. Comparing with results in Tables~\ref{tab:results_evaluation_large_FO} and~\ref{tab:results_evaluation_large_FU}, no significant performance differences were observed. This could be attributed to the substantial disparity between the ImageNet dataset and our UrbanSARFloods dataset. ImageNet comprises millions of RGB neutral images, while UrbanSARFloods consists of 8-band SAR data, including SAR intensity and InSAR coherence data. This difference in feature spaces between the input channels of the source domain (ImageNet) and the target domain (UrbanSARFloods) may impede the models' ability to extract relevant features for flood classification in our study. Similar results were obtained when using manually annotated ground truth data for subsets of the entire Sentinel-1 image frame. Hence, further exploration of advanced transfer learning techniques, such as domain adaptation, is essential to address challenges in large-scale urban flood mapping.

\begin{table*}[!]
\caption{Evaluation of results using semi-supervised labels across the entire Sentinel-1 image frame for Flooded open areas (FO)}
\label{tab:results_evaluation_large_FO}
\centering
\fontsize{8}{10}\selectfont
\scalebox{1.05}{
\begin{tabular}{c|ccc|ccc|ccc}
\hline
\multirow{2}{*}{} & \multicolumn{3}{c|}{Weihui}                                         & \multicolumn{3}{c|}{Jubba}                                          & \multicolumn{3}{c}{NovaKakhovka}                                    \\ \cline{2-10} 
                  & \multicolumn{1}{c}{Precision $\uparrow$} & \multicolumn{1}{c}{Recall $\uparrow$} & F1 $\uparrow$  & \multicolumn{1}{c}{Precision $\uparrow$} & \multicolumn{1}{c}{Recall $\uparrow$} & F1 $\uparrow$  & \multicolumn{1}{c}{Precision $\uparrow$} & \multicolumn{1}{c}{Recall $\uparrow$} & F1 $\uparrow$  \\ \hline
Unet              & \multicolumn{1}{c|}{0.54}      & \multicolumn{1}{c|}{\cellcolor{gray!25}{0.99}}   & 0.70 & \multicolumn{1}{c|}{0.38}      & \multicolumn{1}{c|}{\cellcolor{gray!25}{0.99}}   & 0.55 & \multicolumn{1}{c|}{0.20}      & \multicolumn{1}{c|}{0.81}   & 0.32 \\ \hline
Unet++            & \multicolumn{1}{c|}{\cellcolor{gray!25}{0.64}}      & \multicolumn{1}{c|}{0.98}   & \cellcolor{gray!25}{0.77} & \multicolumn{1}{c|}{\cellcolor{gray!25}{0.50}}      & \multicolumn{1}{c|}{\cellcolor{gray!25}{0.99}}   & \cellcolor{gray!25}{0.66} & \multicolumn{1}{c|}{\cellcolor{gray!25}{0.22}}      & \multicolumn{1}{c|}{0.89}   & \cellcolor{gray!25}{0.35} \\ \hline
MANet             & \multicolumn{1}{c|}{0.50}      & \multicolumn{1}{c|}{0.98}   & 0.67 & \multicolumn{1}{c|}{0.45}      & \multicolumn{1}{c|}{\cellcolor{gray!25}{0.99}}   & 0.62 & \multicolumn{1}{c|}{0.12}      & \multicolumn{1}{c|}{\cellcolor{gray!25}{0.96}}   & 0.21 \\ \hline
Linknet           & \multicolumn{1}{c|}{0.46}      & \multicolumn{1}{c|}{\cellcolor{gray!25}{0.99}}   & 0.63 & \multicolumn{1}{c|}{0.46}      & \multicolumn{1}{c|}{\cellcolor{gray!25}{0.99}}   & 0.63 & \multicolumn{1}{c|}{0.11}      & \multicolumn{1}{c|}{0.93}   & 0.20 \\ \hline
FPN               & \multicolumn{1}{c|}{0.56}      & \multicolumn{1}{c|}{0.97}   & 0.71 & \multicolumn{1}{c|}{0.46}      & \multicolumn{1}{c|}{\cellcolor{gray!25}{0.99}}   & 0.63 & \multicolumn{1}{c|}{0.19}      & \multicolumn{1}{c|}{0.86}   & 0.31 \\ \hline
PSPNet            & \multicolumn{1}{c|}{0.40}      & \multicolumn{1}{c|}{0.97}   & 0.57 & \multicolumn{1}{c|}{0.34}      & \multicolumn{1}{c|}{\cellcolor{gray!25}{0.99}}   & 0.51 & \multicolumn{1}{c|}{0.12}      & \multicolumn{1}{c|}{0.90}   & 0.22 \\ \hline
PAN               & \multicolumn{1}{c|}{0.46}      & \multicolumn{1}{c|}{0.98}   & 0.62 & \multicolumn{1}{c|}{0.37}      & \multicolumn{1}{c|}{\cellcolor{gray!25}{0.99}}   & 0.54 & \multicolumn{1}{c|}{0.14}      & \multicolumn{1}{c|}{0.95}   & 0.24 \\ \hline
Deeplab v3        & \multicolumn{1}{c|}{0.49}      & \multicolumn{1}{c|}{0.98}   & 0.65 & \multicolumn{1}{c|}{0.34}      & \multicolumn{1}{c|}{\cellcolor{gray!25}{0.99}}   & 0.51 & \multicolumn{1}{c|}{0.15}      & \multicolumn{1}{c|}{0.95}   & 0.26 \\ \hline
Deeplab v3   +    & \multicolumn{1}{c|}{0.56}      & \multicolumn{1}{c|}{0.98}   & 0.72 & \multicolumn{1}{c|}{0.42}      & \multicolumn{1}{c|}{\cellcolor{gray!25}{0.99}}   & 0.59 & \multicolumn{1}{c|}{0.18}      & \multicolumn{1}{c|}{0.88}   & 0.30 \\ \hline
\end{tabular}}
\end{table*}

\begin{table*}[!]
\caption{Evaluation of results using semi-supervised labels across the entire Sentinel-1 image frame for Flooded urban areas (FU)}
\label{tab:results_evaluation_large_FU}
\centering
\fontsize{8}{10}\selectfont
\begin{tabular}{c|ccc|ccc|ccc}
\hline
\multirow{2}{*}{} & \multicolumn{3}{c|}{Weihui}                                         & \multicolumn{3}{c|}{Jubba}                                          & \multicolumn{3}{c}{NovaKakhovka}                                    \\ \cline{2-10} 
                  & \multicolumn{1}{c}{Precision $\uparrow$} & \multicolumn{1}{c}{Recall $\uparrow$} & F1 $\uparrow$   & \multicolumn{1}{c}{Precision $\uparrow$} & \multicolumn{1}{c}{Recall $\uparrow$} & F1 $\uparrow$  & \multicolumn{1}{c}{Precision $\uparrow$} & \multicolumn{1}{c}{Recall $\uparrow$} & F1 $\uparrow$  \\ \hline
Unet              & \multicolumn{1}{c|}{\cellcolor{gray!25}{0.06}}      & \multicolumn{1}{c|}{0.32}   & \cellcolor{gray!25}{0.10} & \multicolumn{1}{c|}{\cellcolor{gray!25}{0.37}}      & \multicolumn{1}{c|}{0.79}   & \cellcolor{gray!25}{0.50} & \multicolumn{1}{c|}{0.02}      & \multicolumn{1}{c|}{0.41}   & \cellcolor{gray!25}{0.04} \\ \hline
Unet++            & \multicolumn{1}{c|}{0.01}      & \multicolumn{1}{c|}{0.97}   & 0.02 & \multicolumn{1}{c|}{0.12}      & \multicolumn{1}{c|}{\cellcolor{gray!25}{0.95}}   & 0.21 & \multicolumn{1}{c|}{0.01}      & \multicolumn{1}{c|}{\cellcolor{gray!25}{0.86}}   & 0.02 \\ \hline
MANet             & \multicolumn{1}{c|}{0.02}      & \multicolumn{1}{c|}{0.93}   & 0.04 & \multicolumn{1}{c|}{0.19}      & \multicolumn{1}{c|}{0.83}   & 0.31 & \multicolumn{1}{c|}{0.01}      & \multicolumn{1}{c|}{0.61}   & 0.02 \\ \hline
Linknet           & \multicolumn{1}{c|}{0.03}      & \multicolumn{1}{c|}{0.61}   & 0.07 & \multicolumn{1}{c|}{0.07}      & \multicolumn{1}{c|}{0.63}   & 0.12 & \multicolumn{1}{c|}{\cellcolor{gray!25}{0.12}}      & \multicolumn{1}{c|}{0.39}   & 0.03 \\ \hline
FPN               & \multicolumn{1}{c|}{0.01}      & \multicolumn{1}{c|}{0.90}   & 0.01 & \multicolumn{1}{c|}{0.07}      & \multicolumn{1}{c|}{0.90}   & 0.13 & \multicolumn{1}{c|}{0.01}      & \multicolumn{1}{c|}{0.75}   & 0.01 \\ \hline
PSPNet            & \multicolumn{1}{c|}{0.01}      & \multicolumn{1}{c|}{0.91}   & 0.02 & \multicolumn{1}{c|}{0.08}      & \multicolumn{1}{c|}{\cellcolor{gray!25}{0.95}}   & 0.15 & \multicolumn{1}{c|}{0.01}      & \multicolumn{1}{c|}{0.81}   & 0.01 \\ \hline
PAN               & \multicolumn{1}{c|}{0.01}      & \multicolumn{1}{c|}{\cellcolor{gray!25}{0.98}}   & 0.01 & \multicolumn{1}{c|}{0.07}      & \multicolumn{1}{c|}{0.90}   & 0.13 & \multicolumn{1}{c|}{0.01}      & \multicolumn{1}{c|}{0.70}   & 0.02 \\ \hline
Deeplab v3        & \multicolumn{1}{c|}{0.01}      & \multicolumn{1}{c|}{0.88}   & 0.01 & \multicolumn{1}{c|}{0.12}      & \multicolumn{1}{c|}{0.74}   & 0.21 & \multicolumn{1}{c|}{0.01}      & \multicolumn{1}{c|}{0.77}   & 0.01 \\ \hline
Deeplab v3   +    & \multicolumn{1}{c|}{0.01}      & \multicolumn{1}{c|}{0.87}   & 0.03 & \multicolumn{1}{c|}{0.13}      & \multicolumn{1}{c|}{0.89}   & 0.23 & \multicolumn{1}{c|}{0.01}      & \multicolumn{1}{c|}{0.77}   & 0.02 \\ \hline
\end{tabular}
\end{table*}

\begin{table*}[!]
\caption{Evaluation of results using semi-supervised labels across the entire Sentinel-1 image frame for Flooded open areas (FO) using the pretrained models}
\label{tab:results_evaluation_large_FO_pretrain}
\centering
\fontsize{8}{10}\selectfont
\begin{tabular}{c|ccc|ccc|ccc}
\hline
\multirow{2}{*}{} & \multicolumn{3}{c|}{Weihui}                                         & \multicolumn{3}{c|}{Jubba}                                          & \multicolumn{3}{c}{NovaKakhovka}                                    \\ \cline{2-10} 
                  & \multicolumn{1}{c}{Precision $\uparrow$} & \multicolumn{1}{c}{Recall $\uparrow$} & F1 $\uparrow$  & \multicolumn{1}{c}{Precision $\uparrow$} & \multicolumn{1}{c}{Recall $\uparrow$} & F1 $\uparrow$   & \multicolumn{1}{c}{Precision $\uparrow$} & \multicolumn{1}{c}{Recall $\uparrow$} & F1 $\uparrow$  \\ \hline
Pretrained   Unet         & \multicolumn{1}{c|}{\cellcolor{gray!25}{0.65}}      & \multicolumn{1}{c|}{0.98}   & \cellcolor{gray!25}{0.78} & \multicolumn{1}{c|}{\cellcolor{gray!25}{0.60}}      & \multicolumn{1}{c|}{0.98}   & \cellcolor{gray!25}{0.74} & \multicolumn{1}{c|}{0.19}      & \multicolumn{1}{c|}{0.88}   & 0.31 \\ \hline
Pretrained   Unet++       & \multicolumn{1}{c|}{0.50}      & \multicolumn{1}{c|}{\cellcolor{gray!25}{0.99}}   & 0.66 & \multicolumn{1}{c|}{0.43}      & \multicolumn{1}{c|}{\cellcolor{gray!25}{0.99}}   & 0.60 & \multicolumn{1}{c|}{0.16}      & \multicolumn{1}{c|}{0.88}   & 0.27 \\ \hline
Pretrained   MANet        & \multicolumn{1}{c|}{0.46}      & \multicolumn{1}{c|}{\cellcolor{gray!25}{0.99}}   & 0.63 & \multicolumn{1}{c|}{0.28}      & \multicolumn{1}{c|}{\cellcolor{gray!25}{\cellcolor{gray!25}{0.99}}}   & 0.44 & \multicolumn{1}{c|}{0.11}      & \multicolumn{1}{c|}{0.94}   & 0.20 \\ \hline
Pretrained   Linknet      & \multicolumn{1}{c|}{0.53}      & \multicolumn{1}{c|}{\cellcolor{gray!25}{0.99}}   & 0.69 & \multicolumn{1}{c|}{0.35}      & \multicolumn{1}{c|}{\cellcolor{gray!25}{0.99}}   & 0.52 & \multicolumn{1}{c|}{0.12}      & \multicolumn{1}{c|}{0.94}   & 0.21 \\ \hline
Pretrained FPN            & \multicolumn{1}{c|}{0.52}      & \multicolumn{1}{c|}{\cellcolor{gray!25}{0.99}}   & 0.69 & \multicolumn{1}{c|}{0.44}      & \multicolumn{1}{c|}{\cellcolor{gray!25}{0.99}}   & 0.61 & \multicolumn{1}{c|}{0.14}      & \multicolumn{1}{c|}{\cellcolor{gray!25}{0.96}}   & 0.25 \\ \hline
Pretrained   PSPNet       & \multicolumn{1}{c|}{0.58}      & \multicolumn{1}{c|}{0.94}   & 0.71 & \multicolumn{1}{c|}{0.50}      & \multicolumn{1}{c|}{0.98}   & 0.66 & \multicolumn{1}{c|}{\cellcolor{gray!25}{0.30}}      & \multicolumn{1}{c|}{0.74}   & \cellcolor{gray!25}{0.43} \\ \hline
Pretrained PAN            & \multicolumn{1}{c|}{0.44}      & \multicolumn{1}{c|}{0.98}   & 0.61 & \multicolumn{1}{c|}{0.47}      & \multicolumn{1}{c|}{\cellcolor{gray!25}{0.99}}   & 0.64 & \multicolumn{1}{c|}{0.21}      & \multicolumn{1}{c|}{0.84}   & 0.34 \\ \hline
Pretrained   Deeplab v3   & \multicolumn{1}{c|}{0.44}      & \multicolumn{1}{c|}{0.95}   & 0.60 & \multicolumn{1}{c|}{0.33}      & \multicolumn{1}{c|}{\cellcolor{gray!25}{0.99}}   & 0.49 & \multicolumn{1}{c|}{0.10}      & \multicolumn{1}{c|}{0.92}   & 0.18 \\ \hline
Pretrained   Deeplab v3 + & \multicolumn{1}{c|}{0.40}      & \multicolumn{1}{c|}{\cellcolor{gray!25}{0.99}}   & 0.57 & \multicolumn{1}{c|}{0.29}      & \multicolumn{1}{c|}{\cellcolor{gray!25}{0.99}}   & 0.45 & \multicolumn{1}{c|}{0.09}      & \multicolumn{1}{c|}{0.94}   & 0.17 \\ \hline
\end{tabular}
\end{table*}

\begin{table*}[!]
\caption{Evaluation of results using semi-supervised labels across the entire Sentinel-1 image frame for Flooded urban areas (FU) using the pretrained models}
\label{tab:results_evaluation_large_FU_pretrain}
\centering
\fontsize{8}{10}\selectfont
\begin{tabular}{c|ccc|ccc|ccc}
\hline
\multirow{2}{*}{} & \multicolumn{3}{c|}{Weihui}                                         & \multicolumn{3}{c|}{Jubba}                                          & \multicolumn{3}{c}{NovaKakhovka}                                    \\ \cline{2-10} 
                  & \multicolumn{1}{c}{Precision $\uparrow$} & \multicolumn{1}{c}{Recall $\uparrow$} & F1 $\uparrow$   & \multicolumn{1}{c}{Precision $\uparrow$} & \multicolumn{1}{c}{Recall $\uparrow$} & F1 $\uparrow$   & \multicolumn{1}{c}{Precision $\uparrow$} & \multicolumn{1}{c}{Recall $\uparrow$} & F1 $\uparrow$   \\ \hline
Pretrained   Unet         & \multicolumn{1}{c|}{0.02}      & \multicolumn{1}{c|}{0.74}   & 0.05 & \multicolumn{1}{c|}{0.15}      & \multicolumn{1}{c|}{0.87}   & 0.25 & \multicolumn{1}{c|}{0.01}      & \multicolumn{1}{c|}{0.70}   & 0.02 \\ \hline
Pretrained   Unet++       & \multicolumn{1}{c|}{\cellcolor{gray!25}{0.04}}      & \multicolumn{1}{c|}{0.57}   & \cellcolor{gray!25}{0.07} & \multicolumn{1}{c|}{\cellcolor{gray!25}{0.20}}      & \multicolumn{1}{c|}{0.79}   & \cellcolor{gray!25}{0.32} & \multicolumn{1}{c|}{\cellcolor{gray!25}{0.03}}      & \multicolumn{1}{c|}{0.63}   & \cellcolor{gray!25}{0.05} \\ \hline
Pretrained   MANet        & \multicolumn{1}{c|}{0.01}      & \multicolumn{1}{c|}{0.97}   & 0.02 & \multicolumn{1}{c|}{0.10}      & \multicolumn{1}{c|}{\cellcolor{gray!25}{0.94}}   & 0.19 & \multicolumn{1}{c|}{0.01}      & \multicolumn{1}{c|}{0.73}   & 0.01 \\ \hline
Pretrained   Linknet      & \multicolumn{1}{c|}{0.01}      & \multicolumn{1}{c|}{0.96}   & 0.03 & \multicolumn{1}{c|}{0.08}      & \multicolumn{1}{c|}{0.87}   & 0.15 & \multicolumn{1}{c|}{0.01}      & \multicolumn{1}{c|}{0.59}   & 0.02 \\ \hline
Pretrained FPN            & \multicolumn{1}{c|}{0.02}      & \multicolumn{1}{c|}{0.85}   & 0.04 & \multicolumn{1}{c|}{0.15}      & \multicolumn{1}{c|}{0.72}   & 0.25 & \multicolumn{1}{c|}{0.01}       & \multicolumn{1}{c|}{0.50}   & 0.02 \\ \hline
Pretrained   PSPNet       & \multicolumn{1}{c|}{0.01}      & \multicolumn{1}{c|}{0.97}   & 0.01 & \multicolumn{1}{c|}{0.07}      & \multicolumn{1}{c|}{0.93}   & 0.12 & \multicolumn{1}{c|}{0.01}      & \multicolumn{1}{c|}{\cellcolor{gray!25}{0.92}}   & 0.01 \\ \hline
Pretrained PAN            & \multicolumn{1}{c|}{0.01}      & \multicolumn{1}{c|}{0.83}   & 0.02 & \multicolumn{1}{c|}{0.11}      & \multicolumn{1}{c|}{0.92}   & 0.20 & \multicolumn{1}{c|}{0.01}      & \multicolumn{1}{c|}{0.61}   & 0.03 \\ \hline
Pretrained   Deeplab v3   & \multicolumn{1}{c|}{0.01}      & \multicolumn{1}{c|}{\cellcolor{gray!25}{0.98}}   & 0.01 & \multicolumn{1}{c|}{0.07}      & \multicolumn{1}{c|}{0.92}   & 0.13 & \multicolumn{1}{c|}{0.01}      & \multicolumn{1}{c|}{0.67}   & 0.01 \\ \hline
Pretrained   Deeplab v3 + & \multicolumn{1}{c|}{0.02}      & \multicolumn{1}{c|}{0.69}   & 0.04 & \multicolumn{1}{c|}{0.07}      & \multicolumn{1}{c|}{0.70}   & 0.13 & \multicolumn{1}{c|}{0.02}      & \multicolumn{1}{c|}{0.51}   & 0.04 \\ \hline
\end{tabular}
\end{table*}
% \section{Discussion}
\section{Conclusion}

One of the bottlenecks in integrating DL techniques with large-scale urban flood mapping is the lack of proper open-access datasets. To address this gap, we constructed a pre-processed Sentinel-1 dataset, known as UrbanSARFloods, encompassing both urban and rural floods. This dataset encapsulates two significant challenges encountered in large-scale remote sensing mapping: complex background samples and imbalanced data. We evaluated state-of-the-art methods on the UrbanSARFloods dataset, uncovering the specific challenges posed by UrbanSARFloods. Additionally, we conducted transfer learning experiments to explore alternative approaches for overcoming these challenges. 

This work offers a free and open dataset to advance large-scale urban flood mapping in the area of microwave remote sensing. We also provide this benchmarked task with two considerable challenges, allowing other researchers to easily build on this work and create new and enhanced capabilities. A potential positive societal impact may arise from the development of generalizable models that can produce large-scale flood maps considering urban floods and rural floods accurately. This could help provide global-scale flood maps using all achieved satellite data.
% \section{Limitation and future work}

\section*{Acknowledgement}

This work is jointly supported by German Federal Ministry for Economic Affairs and Climate Action in the framework of the "national center of excellence ML4Earth" (grant number: 50EE2201C) and by the German Federal Ministry of Education and Research (BMBF) in the framework of the international future AI lab "AI4EO -- Artificial Intelligence for Earth Observation: Reasoning, Uncertainties, Ethics and Beyond" (grant number: 01DD20001).

{
    \small
    % \clearpage
    % \newpage 
    \bibliographystyle{ieeenat_fullname}
    \bibliography{main}

\begin{thebibliography}{54}
\providecommand{\natexlab}[1]{#1}
\providecommand{\url}[1]{\texttt{#1}}
\expandafter\ifx\csname urlstyle\endcsname\relax
  \providecommand{\doi}[1]{doi: #1}\else
  \providecommand{\doi}{doi: \begingroup \urlstyle{rm}\Url}\fi

\bibitem[Badrinarayanan et~al.(2017)Badrinarayanan, Kendall, and Cipolla]{badrinarayanan2017segnet}
Vijay Badrinarayanan, Alex Kendall, and Roberto Cipolla.
\newblock Segnet: A deep convolutional encoder-decoder architecture for image segmentation.
\newblock \emph{IEEE transactions on pattern analysis and machine intelligence}, 39\penalty0 (12):\penalty0 2481--2495, 2017.

\bibitem[Baghermanesh et~al.(2022)Baghermanesh, Jabari, and McGrath]{baghermanesh2022urban}
Shadi~Sadat Baghermanesh, Shabnam Jabari, and Heather McGrath.
\newblock Urban flood detection using {TerraSAR-X and SAR Simulated Reflectivity Maps}.
\newblock \emph{Remote Sensing}, 14\penalty0 (23):\penalty0 6154, 2022.

\bibitem[Barz et~al.(2019)Barz, Schr{\"o}ter, M{\"u}nch, Yang, Unger, Dransch, and Denzler]{barz2019enhancing}
Bj{\"o}rn Barz, Kai Schr{\"o}ter, Moritz M{\"u}nch, Bin Yang, Andrea Unger, Doris Dransch, and Joachim Denzler.
\newblock Enhancing flood impact analysis using interactive retrieval of social media images.
\newblock \emph{arXiv preprint arXiv:1908.03361}, 2019.

\bibitem[Bauer-Marschallinger et~al.(2022)Bauer-Marschallinger, Cao, Tupas, Roth, Navacchi, Melzer, Freeman, and Wagner]{bauer2022satellite}
Bernhard Bauer-Marschallinger, Senmao Cao, Mark~Edwin Tupas, Florian Roth, Claudio Navacchi, Thomas Melzer, Vahid Freeman, and Wolfgang Wagner.
\newblock Satellite-based flood mapping through bayesian inference from a sentinel-1 sar datacube.
\newblock \emph{Remote Sensing}, 14\penalty0 (15):\penalty0 3673, 2022.

\bibitem[Bonafilia et~al.(2020)Bonafilia, Tellman, Anderson, and Issenberg]{bonafilia2020sen1floods11}
Derrick Bonafilia, Beth Tellman, Tyler Anderson, and Erica Issenberg.
\newblock Sen1floods11: A georeferenced dataset to train and test deep learning flood algorithms for sentinel-1.
\newblock In \emph{Proceedings of the IEEE/CVF Conference on Computer Vision and Pattern Recognition Workshops}, pages 210--211, 2020.

\bibitem[Brill et~al.(2021)Brill, Schlaffer, Martinis, Schr{\"o}ter, and Kreibich]{brill2021extrapolating}
Fabio Brill, Stefan Schlaffer, Sandro Martinis, Kai Schr{\"o}ter, and Heidi Kreibich.
\newblock Extrapolating satellite-based flood masks by one-class classification—a test case in houston.
\newblock \emph{Remote Sensing}, 13\penalty0 (11):\penalty0 2042, 2021.

\bibitem[Buchhorn et~al.(2020)Buchhorn, Smets, Bertels, De~Roo, Lesiv, Tsendbazar, Herold, and Fritz]{buchhorn2020copernicus}
Marcel Buchhorn, Bruno Smets, Luc Bertels, Bert De~Roo, Myroslava Lesiv, Nandin-Erdene Tsendbazar, Martin Herold, and Steffen Fritz.
\newblock Copernicus global land service: Land cover 100m: collection 3: epoch 2019: Globe.
\newblock \emph{Version V3. 0.1}, 2020.

\bibitem[Bullock et~al.(2019)Bullock, Cuesta-L{\'a}zaro, and Quera-Bofarull]{bullock2019xnet}
Joseph Bullock, Carolina Cuesta-L{\'a}zaro, and Arnau Quera-Bofarull.
\newblock Xnet: a convolutional neural network (cnn) implementation for medical x-ray image segmentation suitable for small datasets.
\newblock In \emph{Medical Imaging 2019: Biomedical Applications in Molecular, Structural, and Functional Imaging}, pages 453--463. SPIE, 2019.

\bibitem[Caretta et~al.(2022)Caretta, Mukherji, Arfanuzzaman, Betts, Gelfan, Hirabayashi, Lissner, Liu, Gunn, Morgan, Mwanga, Supratid, Pörtner, Roberts, Tignor, Poloczanska, Mintenbeck, Alegría, Craig, Langsdorf, Löschke, Möller, Okem, and Rama]{Caretta2022water}
M.A. Caretta, A. Mukherji, M. Arfanuzzaman, R.A. Betts, A. Gelfan, Y. Hirabayashi, T.K. Lissner, J. Liu, E.L. Gunn, R. Morgan, S. Mwanga, S. Supratid, H.-O. Pörtner, D.C. Roberts, M. Tignor, E.S. Poloczanska, K. Mintenbeck, A. Alegría, M. Craig, S. Langsdorf, S. Löschke, V. Möller, A. Okem, and B. Rama.
\newblock Water. in: Climate change 2022: Impacts, adaptation and vulnerability. contribution of working group ii to the sixth assessment report of the intergovernmental panel on climate change.
\newblock Technical report, Cambridge University, 2022.

\bibitem[Chaurasia and Culurciello(2017)]{chaurasia2017linknet}
Abhishek Chaurasia and Eugenio Culurciello.
\newblock Linknet: Exploiting encoder representations for efficient semantic segmentation.
\newblock In \emph{2017 IEEE visual communications and image processing (VCIP)}, pages 1--4. IEEE, 2017.

\bibitem[Chen et~al.(2017)Chen, Papandreou, Schroff, and Adam]{chen2017rethinking}
Liang-Chieh Chen, George Papandreou, Florian Schroff, and Hartwig Adam.
\newblock Rethinking atrous convolution for semantic image segmentation.
\newblock \emph{arXiv preprint arXiv:1706.05587}, 2017.

\bibitem[Chen et~al.(2018)Chen, Zhu, Papandreou, Schroff, and Adam]{chen2018encoder}
Liang-Chieh Chen, Yukun Zhu, George Papandreou, Florian Schroff, and Hartwig Adam.
\newblock Encoder-decoder with atrous separable convolution for semantic image segmentation.
\newblock In \emph{Proceedings of the European conference on computer vision (ECCV)}, pages 801--818, 2018.

\bibitem[Chini et~al.(2017)Chini, Hostache, Giustarini, and Matgen]{chini2017hierarchical}
Marco Chini, Renaud Hostache, Laura Giustarini, and Patrick Matgen.
\newblock A hierarchical split-based approach for parametric thresholding of sar images: Flood inundation as a test case.
\newblock \emph{IEEE Transactions on Geoscience and Remote Sensing}, 55\penalty0 (12):\penalty0 6975--6988, 2017.

\bibitem[Chini et~al.(2019)Chini, Pelich, Pulvirenti, Pierdicca, Hostache, and Matgen]{chini2019sentinel}
Marco Chini, Ramona Pelich, Luca Pulvirenti, Nazzareno Pierdicca, Renaud Hostache, and Patrick Matgen.
\newblock Sentinel-1 insar coherence to detect floodwater in urban areas: Houston and hurricane harvey as a test case.
\newblock \emph{Remote Sensing}, 11\penalty0 (2):\penalty0 107, 2019.

\bibitem[Fan et~al.(2020)Fan, Wang, Li, and Wang]{fan2020ma}
Tongle Fan, Guanglei Wang, Yan Li, and Hongrui Wang.
\newblock Ma-net: A multi-scale attention network for liver and tumor segmentation.
\newblock \emph{IEEE Access}, 8:\penalty0 179656--179665, 2020.

\bibitem[Garg et~al.(2023)Garg, Feinstein, Timnat, Batchu, Dror, Rosenthal, and Gulshan]{garg2023cross}
Shubhika Garg, Ben Feinstein, Shahar Timnat, Vishal Batchu, Gideon Dror, Adi~Gerzi Rosenthal, and Varun Gulshan.
\newblock Cross modal distillation for flood extent mapping.
\newblock \emph{arXiv preprint arXiv:2302.08180}, 2023.

\bibitem[Gokon et~al.(2023)Gokon, Endo, and Koshimura]{gokon2023detecting}
Hideomi Gokon, Fuyuki Endo, and Shunichi Koshimura.
\newblock Detecting urban floods with small and large scale analysis of alos-2/palsar-2 data.
\newblock \emph{Remote Sensing}, 15\penalty0 (2):\penalty0 532, 2023.

\bibitem[Iakubovskii(2019)]{Iakubovskii:2019}
Pavel Iakubovskii.
\newblock Segmentation models pytorch.
\newblock \url{https://github.com/qubvel/segmentation_models.pytorch}, 2019.

\bibitem[Iervolino et~al.(2015)Iervolino, Guida, Iodice, and Riccio]{iervolino2015flooding}
Pasquale Iervolino, Raffaella Guida, Antonio Iodice, and Daniele Riccio.
\newblock Flooding water depth estimation with high-resolution {SAR}.
\newblock \emph{IEEE Transactions on Geoscience and Remote Sensing}, 53\penalty0 (5):\penalty0 2295--2307, 2015.

\bibitem[Jamali et~al.(2024)Jamali, Roy, Beni, Pradhan, Li, and Ghamisi]{jamali2024residual}
Ali Jamali, Swalpa~Kumar Roy, Leila~Hashemi Beni, Biswajeet Pradhan, Jonathan Li, and Pedram Ghamisi.
\newblock Residual wave vision u-net for flood mapping using dual polarization sentinel-1 sar imagery.
\newblock \emph{International Journal of Applied Earth Observation and Geoinformation}, 127:\penalty0 103662, 2024.

\bibitem[Jiang et~al.(2021)Jiang, Liang, He, Ziegler, Lin, Pan, Wang, Zou, Hao, Mao, et~al.]{jiang2021rapid}
Xin Jiang, Shijing Liang, Xinyue He, Alan~D Ziegler, Peirong Lin, Ming Pan, Dashan Wang, Junyu Zou, Dalei Hao, Ganquan Mao, et~al.
\newblock Rapid and large-scale mapping of flood inundation via integrating spaceborne synthetic aperture radar imagery with unsupervised deep learning.
\newblock \emph{ISPRS journal of photogrammetry and remote sensing}, 178:\penalty0 36--50, 2021.

\bibitem[Karim et~al.(2023)Karim, Sharma, and Barman]{Flood_Area_Segmentation_dataset}
Md~Faizal Karim, Krish Sharma, and Niyar~R Barman.
\newblock Flood area segmentation, 2023.
\newblock Accessed on [8 March 2024].

\bibitem[Katiyar et~al.(2021)Katiyar, Tamkuan, and Nagai]{katiyar2021near}
Vaibhav Katiyar, Nopphawan Tamkuan, and Masahiko Nagai.
\newblock Near-real-time flood mapping using off-the-shelf models with sar imagery and deep learning.
\newblock \emph{Remote Sensing}, 13\penalty0 (12):\penalty0 2334, 2021.

\bibitem[Konapala et~al.(2021)Konapala, Kumar, and Ahmad]{konapala2021exploring}
Goutam Konapala, Sujay~V Kumar, and Shahryar~Khalique Ahmad.
\newblock Exploring sentinel-1 and sentinel-2 diversity for flood inundation mapping using deep learning.
\newblock \emph{ISPRS Journal of Photogrammetry and Remote Sensing}, 180:\penalty0 163--173, 2021.

\bibitem[Li et~al.(2018)Li, Xiong, An, and Wang]{li2018pyramid}
Hanchao Li, Pengfei Xiong, Jie An, and Lingxue Wang.
\newblock Pyramid attention network for semantic segmentation.
\newblock \emph{arXiv preprint arXiv:1805.10180}, 2018.

\bibitem[Li et~al.(2023{\natexlab{a}})Li, Li, Song, Chen, Wang, Bao, Zhang, and Meng]{li2023robust}
Junjie Li, Linyi Li, Yanjiao Song, Jiaming Chen, Zhe Wang, Yi Bao, Wen Zhang, and Lingkui Meng.
\newblock A robust large-scale surface water mapping framework with high spatiotemporal resolution based on the fusion of multi-source remote sensing data.
\newblock \emph{International Journal of Applied Earth Observation and Geoinformation}, 118:\penalty0 103288, 2023{\natexlab{a}}.

\bibitem[Li et~al.(2023{\natexlab{b}})Li, Lee, Wang, Hsu, and Arundel]{li2023assessment}
Wenwen Li, Hyunho Lee, Sizhe Wang, Chia-Yu Hsu, and Samantha~T Arundel.
\newblock Assessment of a new geoai foundation model for flood inundation mapping.
\newblock In \emph{Proceedings of the 6th ACM SIGSPATIAL International Workshop on AI for Geographic Knowledge Discovery}, pages 102--109, 2023{\natexlab{b}}.

\bibitem[Li et~al.(2019)Li, Martinis, and Wieland]{li2019urban}
Yu Li, Sandro Martinis, and Marc Wieland.
\newblock Urban flood mapping with an active self-learning convolutional neural network based on terrasar-x intensity and interferometric coherence.
\newblock \emph{ISPRS Journal of Photogrammetry and Remote Sensing}, 152:\penalty0 178--191, 2019.

\bibitem[Lin et~al.(2017)Lin, Doll{\'a}r, Girshick, He, Hariharan, and Belongie]{lin2017feature}
Tsung-Yi Lin, Piotr Doll{\'a}r, Ross Girshick, Kaiming He, Bharath Hariharan, and Serge Belongie.
\newblock Feature pyramid networks for object detection.
\newblock In \emph{Proceedings of the IEEE conference on computer vision and pattern recognition}, pages 2117--2125, 2017.

\bibitem[Luppino et~al.(2019)Luppino, Bianchi, Moser, and Anfinsen]{8798991}
Luigi~Tommaso Luppino, Filippo~Maria Bianchi, Gabriele Moser, and Stian~Normann Anfinsen.
\newblock Unsupervised image regression for heterogeneous change detection.
\newblock \emph{IEEE Transactions on Geoscience and Remote Sensing}, 57\penalty0 (12):\penalty0 9960--9975, 2019.

\bibitem[Lv et~al.(2022)Lv, Meng, Edwing, Xue, Geng, and Yan]{lv2022high}
Suna Lv, Lingsheng Meng, Deanna Edwing, Sihan Xue, Xupu Geng, and Xiao-Hai Yan.
\newblock High-performance segmentation for flood mapping of hisea-1 sar remote sensing images.
\newblock \emph{Remote Sensing}, 14\penalty0 (21):\penalty0 5504, 2022.

\bibitem[Marconcini et~al.(2021)Marconcini, Metz-Marconcini, Esch, and Gorelick]{marconcini2021understanding}
Mattia Marconcini, Annekatrin Metz-Marconcini, Thomas Esch, and Noel Gorelick.
\newblock Understanding current trends in global urbanisation-the world settlement footprint suite.
\newblock \emph{GI\_Forum}, 9\penalty0 (1):\penalty0 33--38, 2021.

\bibitem[Mateo-Garcia et~al.(2021)Mateo-Garcia, Veitch-Michaelis, Smith, Oprea, Schumann, Gal, Baydin, and Backes]{mateo2021towards}
Gonzalo Mateo-Garcia, Joshua Veitch-Michaelis, Lewis Smith, Silviu~Vlad Oprea, Guy Schumann, Yarin Gal, At{\i}l{\i}m~G{\"u}ne{\c{s}} Baydin, and Dietmar Backes.
\newblock Towards global flood mapping onboard low cost satellites with machine learning.
\newblock \emph{Scientific reports}, 11\penalty0 (1):\penalty0 7249, 2021.

\bibitem[Montello et~al.(2022)Montello, Arnaudo, and Rossi]{bprf-jf62-22}
Fabio Montello, Edoardo Arnaudo, and Claudio Rossi.
\newblock Mmflood: A multimodal dataset for flood delineation from satellite imagery, 2022.

\bibitem[Nemni et~al.(2020)Nemni, Bullock, Belabbes, and Bromley]{nemni2020fully}
Edoardo Nemni, Joseph Bullock, Samir Belabbes, and Lars Bromley.
\newblock Fully convolutional neural network for rapid flood segmentation in synthetic aperture radar imagery.
\newblock \emph{Remote Sensing}, 12\penalty0 (16):\penalty0 2532, 2020.

\bibitem[Orion29()]{github_repo}
Orion29.
\newblock Satellite image segmentation for flood damage analysis.
\newblock \url{https://github.com/orion29/Satellite-Image-Segmentation-for-Flood-Damage-Analysis}.
\newblock Accessed on [8 March 2024].

\bibitem[Pelich et~al.(2022)Pelich, Chini, Hostache, Matgen, Pulvirenti, and Pierdicca]{pelich2022mapping}
Ramona Pelich, Marco Chini, Renaud Hostache, Patrick Matgen, Luca Pulvirenti, and Nazzareno Pierdicca.
\newblock Mapping floods in urban areas from dual-polarization insar coherence data.
\newblock \emph{IEEE Geoscience and Remote Sensing Letters}, 19:\penalty0 1--5, 2022.

\bibitem[Portalés-Julià et~al.(2023)Portalés-Julià, Mateo-García, Purcell, and Gómez-Chova]{julia_global_2023}
Enrique Portalés-Julià, Gonzalo Mateo-García, Cormac Purcell, and Luis Gómez-Chova.
\newblock Global flood extent segmentation in optical satellite images.
\newblock \emph{Scientific Reports}, 13\penalty0 (1):\penalty0 20316, 2023.

\bibitem[Rahnemoonfar et~al.(2021)Rahnemoonfar, Chowdhury, Sarkar, Varshney, Yari, and Murphy]{9460988}
Maryam Rahnemoonfar, Tashnim Chowdhury, Argho Sarkar, Debvrat Varshney, Masoud Yari, and Robin~Roberson Murphy.
\newblock Floodnet: A high resolution aerial imagery dataset for post flood scene understanding.
\newblock \emph{IEEE Access}, 9:\penalty0 89644--89654, 2021.

\bibitem[Rambour et~al.(2020)Rambour, Audebert, Koeniguer, Le~Saux, Crucianu, and Datcu]{w6xz-s898-20}
Clément Rambour, Nicolas Audebert, Elise Koeniguer, Bertrand Le~Saux, Michel Crucianu, and Mihai Datcu.
\newblock Sen12-flood : a sar and multispectral dataset for flood detection, 2020.

\bibitem[Rentschler et~al.(2022)Rentschler, Salhab, and Jafino]{rentschler2022flood}
Jun Rentschler, Melda Salhab, and Bramka~Arga Jafino.
\newblock Flood exposure and poverty in 188 countries.
\newblock \emph{Nature communications}, 13\penalty0 (1):\penalty0 3527, 2022.

\bibitem[Ronneberger et~al.(2015)Ronneberger, Fischer, and Brox]{ronneberger2015u}
Olaf Ronneberger, Philipp Fischer, and Thomas Brox.
\newblock U-net: Convolutional networks for biomedical image segmentation.
\newblock In \emph{Medical Image Computing and Computer-Assisted Intervention--MICCAI 2015: 18th International Conference, Munich, Germany, October 5-9, 2015, Proceedings, Part III 18}, pages 234--241. Springer, 2015.

\bibitem[Saleh et~al.(2023)Saleh, Weng, Holail, Hao, and Xia]{saleh2023dam}
Tamer Saleh, Xingxing Weng, Shimaa Holail, Chen Hao, and Gui-Song Xia.
\newblock Dam-net: Global flood detection from sar imagery using differential attention metric-based vision transformers.
\newblock \emph{arXiv preprint arXiv:2306.00704}, 2023.

\bibitem[Sazara et~al.(2019)Sazara, Cetin, and Iftekharuddin]{sazara_dataset}
Cem Sazara, Mecit Cetin, and Khan Iftekharuddin.
\newblock Image dataset for roadway flooding.
\newblock Mendeley Data, V1, doi: 10.17632/t395bwcvbw.1, 2019.

\bibitem[Shermeyer et~al.(2020)Shermeyer, Hogan, Brown, Van~Etten, Weir, Pacifici, Hansch, Bastidas, Soenen, Bacastow, et~al.]{shermeyer2020spacenet}
Jacob Shermeyer, Daniel Hogan, Jason Brown, Adam Van~Etten, Nicholas Weir, Fabio Pacifici, Ronny Hansch, Alexei Bastidas, Scott Soenen, Todd Bacastow, et~al.
\newblock Spacenet 6: Multi-sensor all weather mapping dataset.
\newblock In \emph{Proceedings of the IEEE/CVF conference on computer vision and pattern recognition workshops}, pages 196--197, 2020.

\bibitem[Xiong et~al.(2022)Xiong, Zhang, Wang, Shi, and Zhu]{xiong2022earthnets}
Zhitong Xiong, Fahong Zhang, Yi Wang, Yilei Shi, and Xiao~Xiang Zhu.
\newblock Earthnets: Empowering ai in earth observation.
\newblock \emph{arXiv preprint arXiv:2210.04936}, 2022.

\bibitem[Yadav et~al.(2024)Yadav, Nascetti, Azizpour, and Ban]{yadav2024unsupervised}
Ritu Yadav, Andrea Nascetti, Hossein Azizpour, and Yifang Ban.
\newblock Unsupervised flood detection on sar time series using variational autoencoder.
\newblock \emph{International Journal of Applied Earth Observation and Geoinformation}, 126:\penalty0 103635, 2024.

\bibitem[Yang et~al.(2021)Yang, Shen, Anagnostou, Mo, Eggleston, and Kettner]{yang2021high}
Qing Yang, Xinyi Shen, Emmanouil~N Anagnostou, Chongxun Mo, Jack~R Eggleston, and Albert~J Kettner.
\newblock A high-resolution flood inundation archive (2016--the present) from sentinel-1 sar imagery over conus.
\newblock \emph{Bulletin of the American Meteorological Society}, pages 1--40, 2021.

\bibitem[Yang et~al.(2023)Yang, Shen, Zhang, Helfrich, Kellndorfer, and Hao]{yang2023promoting}
Qing Yang, Xinyi Shen, Qingyuan Zhang, Sean Helfrich, Josef~M Kellndorfer, and Wei Hao.
\newblock Promoting sar-based urban flood mapping with adversarial generative network and out of distribution detection.
\newblock In \emph{IGARSS 2023-2023 IEEE International Geoscience and Remote Sensing Symposium}, pages 2336--2338. IEEE, 2023.

\bibitem[Zhao et~al.(2017)Zhao, Shi, Qi, Wang, and Jia]{zhao2017pyramid}
Hengshuang Zhao, Jianping Shi, Xiaojuan Qi, Xiaogang Wang, and Jiaya Jia.
\newblock Pyramid scene parsing network.
\newblock In \emph{Proceedings of the IEEE conference on computer vision and pattern recognition}, pages 2881--2890, 2017.

\bibitem[Zhao et~al.(2021{\natexlab{a}})Zhao, Pelich, Hostache, Matgen, Cao, Wagner, and Chini]{zhao2021deriving}
Jie Zhao, Ramona Pelich, Renaud Hostache, Patrick Matgen, Senmao Cao, Wolfgang Wagner, and Marco Chini.
\newblock Deriving exclusion maps from c-band sar time-series in support of floodwater mapping.
\newblock \emph{Remote Sensing of Environment}, 265:\penalty0 112668, 2021{\natexlab{a}}.

\bibitem[Zhao et~al.(2021{\natexlab{b}})Zhao, Pelich, Hostache, Matgen, Wagner, and Chini]{zhao2021large}
Jie Zhao, Ramona Pelich, Renaud Hostache, Patrick Matgen, Wolfgang Wagner, and Marco Chini.
\newblock A large-scale 2005--2012 flood map record derived from envisat-asar data: United kingdom as a test case.
\newblock \emph{Remote Sensing of Environment}, 256:\penalty0 112338, 2021{\natexlab{b}}.

\bibitem[Zhao et~al.(2022)Zhao, Li, Matgen, Pelich, Hostache, Wagner, and Chini]{zhao2022urban}
Jie Zhao, Yu Li, Patrick Matgen, Ramona Pelich, Renaud Hostache, Wolfgang Wagner, and Marco Chini.
\newblock Urban-aware u-net for large-scale urban flood mapping using multitemporal sentinel-1 intensity and interferometric coherence.
\newblock \emph{IEEE Transactions on Geoscience and Remote Sensing}, 60:\penalty0 1--21, 2022.

\bibitem[Zhou et~al.(2018)Zhou, Rahman~Siddiquee, Tajbakhsh, and Liang]{zhou2018unet++}
Zongwei Zhou, Md~Mahfuzur Rahman~Siddiquee, Nima Tajbakhsh, and Jianming Liang.
\newblock Unet++: A nested u-net architecture for medical image segmentation.
\newblock In \emph{Deep Learning in Medical Image Analysis and Multimodal Learning for Clinical Decision Support: 4th International Workshop, DLMIA 2018, and 8th International Workshop, ML-CDS 2018, Held in Conjunction with MICCAI 2018, Granada, Spain, September 20, 2018, Proceedings 4}, pages 3--11. Springer, 2018.

\end{thebibliography}
}

% \input{X_suppl}

\end{document}